\documentclass{article}

\usepackage{times}
\usepackage{graphicx} 
\usepackage{subfigure} 

\usepackage{natbib}

\usepackage{amsmath}
\usepackage{amssymb}

\usepackage{algorithm}
\usepackage{algorithmic}

\usepackage{hyperref}

\usepackage[accepted]{icml2015}

\icmltitlerunning{Improving neural networks with bunches of neurons modeled by Kumaraswamy units}

\begin{document} 

\twocolumn[
\icmltitle{Improving neural networks with bunches of neurons modeled by Kumaraswamy units: Preliminary study}

\icmlauthor{Jakub M. Tomczak}{jakub.tomczak@pwr.edu.pl}
\icmladdress{Wroc\l aw University of Technology,
            wybrze\.{z}e Wyspia\'{n}skiego 27, 50-370, Wroc\l aw, Poland}

\icmlkeywords{neural network, Kumaraswamy distribution, nonlinear unit}

\vskip 0.3in
]

\begin{abstract} 
Deep neural networks have recently achieved state-of-the-art results in many machine learning problems, e.g., speech recognition or object recognition. Hitherto, work on rectified linear units (ReLU) provides empirical and theoretical evidence on performance increase of neural networks comparing to typically used sigmoid activation function. In this paper, we investigate a new manner of improving neural networks by introducing a bunch of copies of the same neuron modeled by the generalized Kumaraswamy distribution. As a result, we propose novel non-linear activation function which we refer to as \textit{Kumaraswamy unit} which is closely related to ReLU. In the experimental study with MNIST image corpora we evaluate the Kumaraswamy unit applied to single-layer (shallow) neural network and report a significant drop in test classification error and test cross-entropy in comparison to sigmoid unit, ReLU and Noisy ReLU. 
\end{abstract} 


\section{Introduction}
\label{sect:introduction}

Deep neural networks are quickly becoming a crucial element of high performance systems in many domains \citep{BCV:13}, e.g., speech recognition, object recognition, natural language processing, multi-task and domain adaptation. Typical neural networks are based on sigmoid hidden units \cite{B:09}, however, they can suffer from the vanishing gradient problem \citep{BSF:94}. The issue may arise when lower layers of a neural network have gradients nearly $0$ because higher layers are mostly saturated at $0$ or $1$. The vanishing gradients may drastically slow down the optimization procedure and eventually may lead to a poor local minimum.

In order to overcome issues associated with the sigmoid non-linearity it is advocated to utilize other types of hidden units. Recently, deep neural networks with rectified linear units (ReLU) have seen success in different applications, e.g., signal processing \citep{ZRMMYLNSVDH:13}, sentiment analysis \cite{GBB:11}, object recognition \cite{JKRL:09}, image analysis \cite{NH:10}. It has been shown that piece-wise linear units, such as ReLU, can compute highly complex and structured functions \cite{MPCB:14}. The practical success and theoretical results on ReLU have indicated a new direction for research. In \citep{MHN:13} a leaky version of ReLU was proposed. The empirical evaluation on speech recognition task has shown slight improvement in comparison to sigmoid unit and ReLU. Further investigations with parametrized leaky ReLU (called Parametric ReLU) in \citep{HZRS:15} confirmed the presumption that simple ReLU may be to restrictive to learn fully successful representation. Recently, \citet{AHSB:15} went even further and proposed Adaptive Piecewise Linear Units (APLU), ReLU with a piecewise linear part for negative values. In the experiments it was shown that APLU can lead to significant performance increase.

In this work, we propose to improve neural networks by modeling neurons in a new manner. Our idea is to replicate a neuron with the same weights and biases in order to increase the robustness of learning a pattern. Moreover, we take into account the complex structure of single neuron which is represented by an additional parameter. A suitable fashion of modeling such \textit{bunch of neurons} is application of a \textit{generalized Kumaraswamy distribution} (\textsc{Kum-G}) \cite{CC:11, NCO:12}. The Kumaraswamy distribution (\textsc{Kum}) can be seen as an alternative distribution to the Beta distribution \cite{J:09} and \textsc{Kum-G} determines a new class of distribution for given base probability measure. In our case, assuming single neuron in the bunch of neurons is modeled by a sigmoid function, we obtain novel non-linear activation function which we refer to as \textit{Kumaraswamy unit}. For properly chosen parameters of \textsc{Kum-G}, the Kumaraswamy unit behaves similarly to ReLU.

The contribution of the paper is the following: i) we introduce an original non-linear activation function (Kumaraswamy unit) which follows from modeling a bunch of copies of the same neuron using the generalized Kumaraswamy distribution, ii) we provide close relationship between the Kumaraswamy unit and ReLU, iii) we provide an empirical evaluation of a single-layer neural network with the proposed hidden unit applied to MNIST dataset.

\section{Modeling bunch of neurons: Kumaraswamy unit}
\label{sect:kumaraswamy}

\paragraph{Preliminaries} Let us focus on conventional feed-forward neural network with an input (visibles) $\mathbf{v} \in [0,1]^{D\times 1}$ and an output $\mathbf{y} \in \{0,1\}^{K\times 1}$ such that $\sum_{k} y_{k} = 1$. In general, the network consists of $L$ hidden layers, however, we restrict our considerations to one hidden layer for clarity. Therefore, the parameters of the network are $\boldsymbol\theta = \{\mathbf{c}, \mathbf{d}, \mathbf{W}, \mathbf{U}\}$, where $\mathbf{c} \in \mathbb{R}^{M\times 1}$ and $\mathbf{d} \in \mathbb{R}^{K\times 1}$ denote hidden and output biases, respectively, and $\mathbf{W} \in \mathbb{R}^{M \times D}$ are input-to-hidden weights, and $\mathbf{U} \in \mathbb{R}^{K \times M}$ are hidden-to-output weights. The output of the network is modeled by the softmax unit:
\begin{equation}\label{eq:softmax}
p(y_{k}=1|\mathbf{v}, \boldsymbol\theta) = \frac{ \exp( \mathbf{U}_{k \cdot} f(\mathbf{v};\mathbf{c},\mathbf{W}) + d_{k} ) }{ \sum_{l} \exp( \mathbf{U}_{l \cdot} f(\mathbf{v};\mathbf{c},\mathbf{W}) + d_{l} ) }
\end{equation}
where $\mathbf{U}_{i \cdot}$ denotes $i$-th row of the matrix $\mathbf{U}$, and $f(\mathbf{v};\mathbf{c},\mathbf{W}) $ is the $M$-dimensional output of the hidden layer.

The activity of hidden units is modeled by some element-wise non-linear function $f(\cdot)$ . Typical activation function is the sigmoid function:
\begin{equation}\label{eq:sigmoid}
\sigma(x) = \frac{1}{1 + \exp(-x)}.
\end{equation}
Recently, several alternatives to the sigmoid function have been used in numerous applications, such as, rectified linear unit (ReLU) \cite{JKRL:09}:
\begin{equation}\label{eq:relu}
r(x) = \max\{0,x\},
\end{equation}
or noisy rectified linear unit (Noisy ReLU) \cite{NH:10}:
\begin{equation}\label{eq:reluNoisy}
n(x) = \max\{0, x + \mathcal{N}(0,v)\},
\end{equation}
where $\mathcal{N}(\cdot, \cdot)$ is a normal probability density function with zero mean and variance $v$.

\begin{figure}[!htbp]
\vskip 0.2in
\begin{center}
\centerline{\includegraphics[width=1.1\columnwidth]{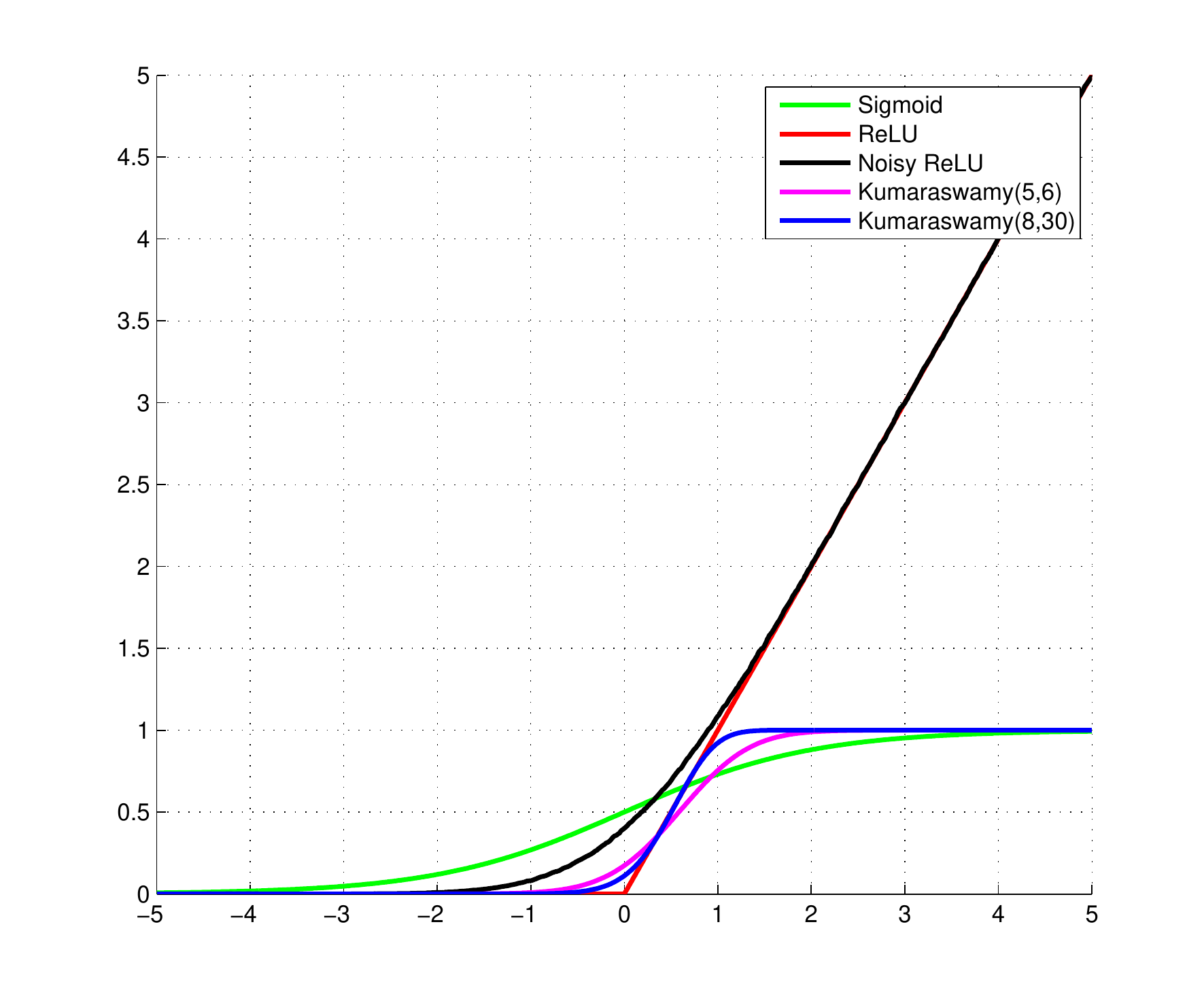}}
\caption{A comparison of chosen non-linear activation functions used in a hidden layer of a neural network. The green curve corresponds to typically used sigmoid function. The red curve represents ReLU non-linearity. The black curve shows the expected value of Noisy ReLU with $v=1$. The magenta and blue curves depict the Kumaraswamy units with different values of scale parameters, $(5,6)$ and $(8,30)$, respectively. }
\label{fig:units}
\end{center}
\vskip -0.2in
\end{figure}

\paragraph{Bunch of neurons} Let us assume that activation of a neuron is modeled by a sigmoid unit. Moreover, let us presume that each neuron consists of $a$ independent elements and there are $b$ independent copies of the same neuron. Therefore, instead of single neuron we consider a bunch of neurons that try to reflect one pattern. A similar idea with replicas of a neuron was introduced in \citep{TH:01} where the replication of sigmoid hidden units with the same weights and biases led to binomial units. Further, it turned out that binomial units with fixed offset to biases resulted in softplus units and its fast approximation, i.e., Noisy ReLU \citep{NH:10}. However, in our approach we copy the sigmoid hidden unit $b$ times and additionally we introduce a second parameter, $a$, which corresponds to modeling complexity of the neuron itself. Increasing the value of $b$ results in higher robustness of the bunch of neurons because it is less probable that the input signal will not activate any hidden unit in the bunch. On the other hand, increasing the value of $a$ leads to higher failure probability of the neuron activation because it suffices that at least one element fails to deactivate the whole neuron.

It turns out that a suitable manner of modeling a bunch of neurons as described above is a \textit{generalized Kumaraswamy distribution} \citep{CC:11, NCO:12}. The generalized Kumaraswamy distribution (\textsc{Kum-G}) for given base distribution $G(x)$ with a probability density function $g(x)$ is defined as follows: 
\begin{equation}\label{eq:kumaraswamyG}
K_{G}(x|a,b) = 1 - (1 - G(x)^{a})^{b},
\end{equation}
where $a>0$ and $b>0$ are shape parameters. The probability density function of $K_{G}(x|a,b)$ has a simple form:
\begin{equation}\label{eq:kumaraswamyGpdf}
k_{G}(x|a,b) = a\ b\ g(x) G(x)^{a-1}(1 -G(x)^{a})^{b-1}.
\end{equation}
For integer-valued shape parameters $a$ and $b$ \textsc{Kum-G} has a nice interpretation of a system which consists of $b$ independent components and each component is made up of $a$ independent subcomponents. \textsc{Kum-G} perfectly fits to modeling chosen property of a complex system, such as, lifetime of an entire system \citep{NCO:12}. We can clearly apply \textsc{Kum-G} to represent the bunch of neurons.

\paragraph{Kumaraswamy unit} Assuming that single neuron activates according to the sigmoid function, we can take advantage of \textsc{Kum-G} to model the bunch of neurons which yields a new kind of non-linear activation function:
\begin{equation}\label{eq:kumaraswamy}
K_{\sigma}(x|a,b) = 1 - (1 - \sigma(x)^{a})^{b}.
\end{equation}
We refer this resulting unit to as \textit{Kumaraswamy unit} (Kumaraswamy$(a,b)$). Obviously, for $a=b=1$ one recovers the sigmoid activation function. In the context of gradient-based learning algorithm it is important to compute a derivative of a hidden unit. Since the derivative of the sigmoid function can be easily calculated, the derivative of the Kumaraswamy unit can be obtained immediately (see Equation \ref{eq:kumaraswamyGpdf}).

An intriguing property of the Kumaraswamy unit is that for properly chosen values of $a$ and $b$ it can behave like ReLU (see Figure \ref{fig:units} for comparison of sigmoid unit, ReLU, Noisy ReLu and the Kumaraswamy unit). We consider only two pairs of values of shape parameters, namely, $(a,b) \in \{(5,6), (8,30)\}$. The Kumaraswamy unit with $a=5$ and $b=6$ is the closest\footnote{In the Euclidean sense.} approximation of ReLU for value $0.5$, while the second pair of values gives the closest approximation of ReLU in points $0.25$ and $0.75$. As we can evidently notice in Figure \ref{fig:units}, the Kumaraswamy unit can behave similarly to ReLU but it returns values between $0$ and $1$ like the sigmoid function. We argue that such behavior may be crucial in training a neural network and is more biologically plausible.

\paragraph{Training} Learning the parameters of the network $\boldsymbol\theta$ for given data $\mathcal{D} = \{(\mathbf{v}_{n}, \mathbf{t}_{n})\}_{n=1}^{N}$ is performed by minimizing the cross-entropy loss. In the case of the neural network with output given by the softmax unit, the cross-entropy loss is equivalent to the negative conditional log-likelihood function:
\begin{equation}\label{eq:crossEntropyError}
\ell(\boldsymbol\theta) = - \sum_{n} \sum_{k} t_{kn} \log p(y_{kn}|\mathbf{v}_{n},\boldsymbol\theta).
\end{equation}

\section{Experiments}
\label{sect:experiments}

\paragraph{Goal} In the experiment we aim at answering the following question:
\begin{itemize}
\item Is the Kumaraswamy unit preferable to sigmoid unit, ReLU or Noisy ReLU in training a single-layer neural network using stochastic gradient descent?
\end{itemize}
We want to point out that we try to verify only the impact of the proposed non-linearity on learning the neural network. We believe that positive answer to the stated question will give us a good starting point for further experiments with multi-layered neural networks, unsupervised pre-training and more sophisticated training techniques.

\paragraph{Data} In the experiment we deal with the well-known MNIST dataset\footnote{http://yann.lecun.com/exdb/mnist/} for hand-written digit classification. We split the dataset to 50,000 training images, 10,000 objects for validation, and 10,000 test examples. Each image consists of $28\times 28$ pixels ($D=784$), and is labeled as one of ten possible digits ($K=10$).

\paragraph{Learning methodology} In the experiment we focus on a single-layer neural network with $500$ hidden units ($M=500$). We train the model using stochastic gradient descent with momentum and a mini-batch size of $100$ examples. The initial value of the momentum term is set according to the model selection with possible values in $\{0, 0.5\}$, and after $50$ epochs it is set to $0.9$. The learning rate is determined in the model selection procedure with possible values in $\{0.001, 0.01, 0.1\}$. During learning process, we apply the learning step policy in which the learning step is divided by $2$ after each $10$ epochs. Additionally, we add a weight decay to the learning objective and we explore the following values of the regularization coefficient: $\{0, 10^{-5}, 10^{-4}\}$. The maximum number of epochs is set to $100$. The number of iterations over the training set is determined using early stopping according to the validation classification error, with a look ahead of $10$ epochs. For all activation units the same initialization of the parameters is applied.

\paragraph{Evaluation methodology} In order to verify whether the Kumaraswamy unit is preferable to the sigmoid unit, ReLU or Noisy ReLU, we use two evaluation metrics, namely, the test classification error (\textbf{Error}) and the test cross-entropy loss (\textbf{Cross-Entropy}).\footnote{In the experiment we report the test cross-entropy loss divided by the number of test examples.} Additionally, we measure the mean activity of hidden units.

\begin{figure}[!htbp]
\vskip 0.2in
\begin{center}
\centerline{\includegraphics[width=1.1\columnwidth]{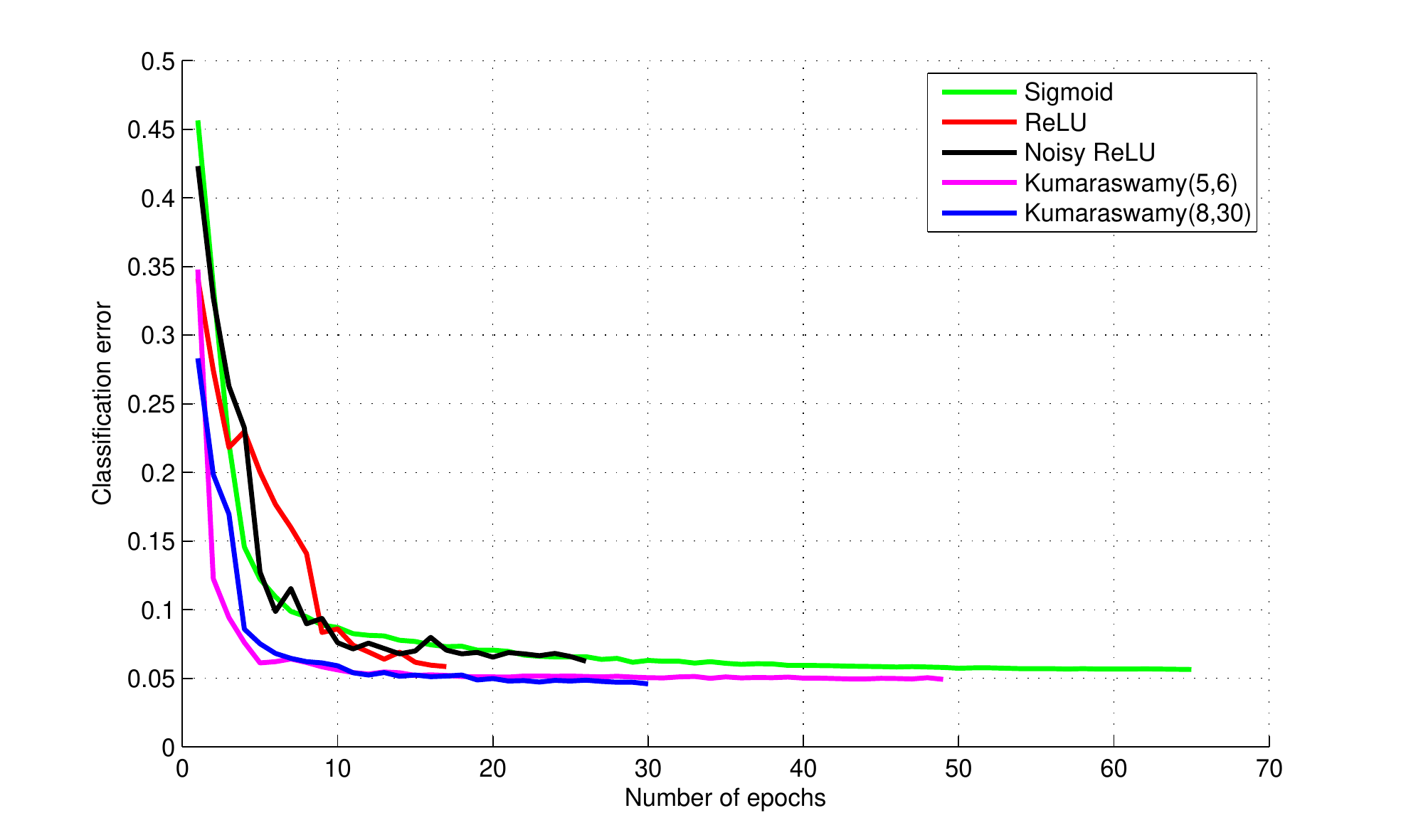}}
\caption{Validation classification error against the number of epochs during learning process for the considered types of hidden units.}
\label{fig:performance}
\end{center}
\vskip -0.2in
\end{figure}

\begin{figure*}[!htbp]
\vskip 0.2in
\begin{center}
\centerline{\includegraphics[width=1\textwidth]{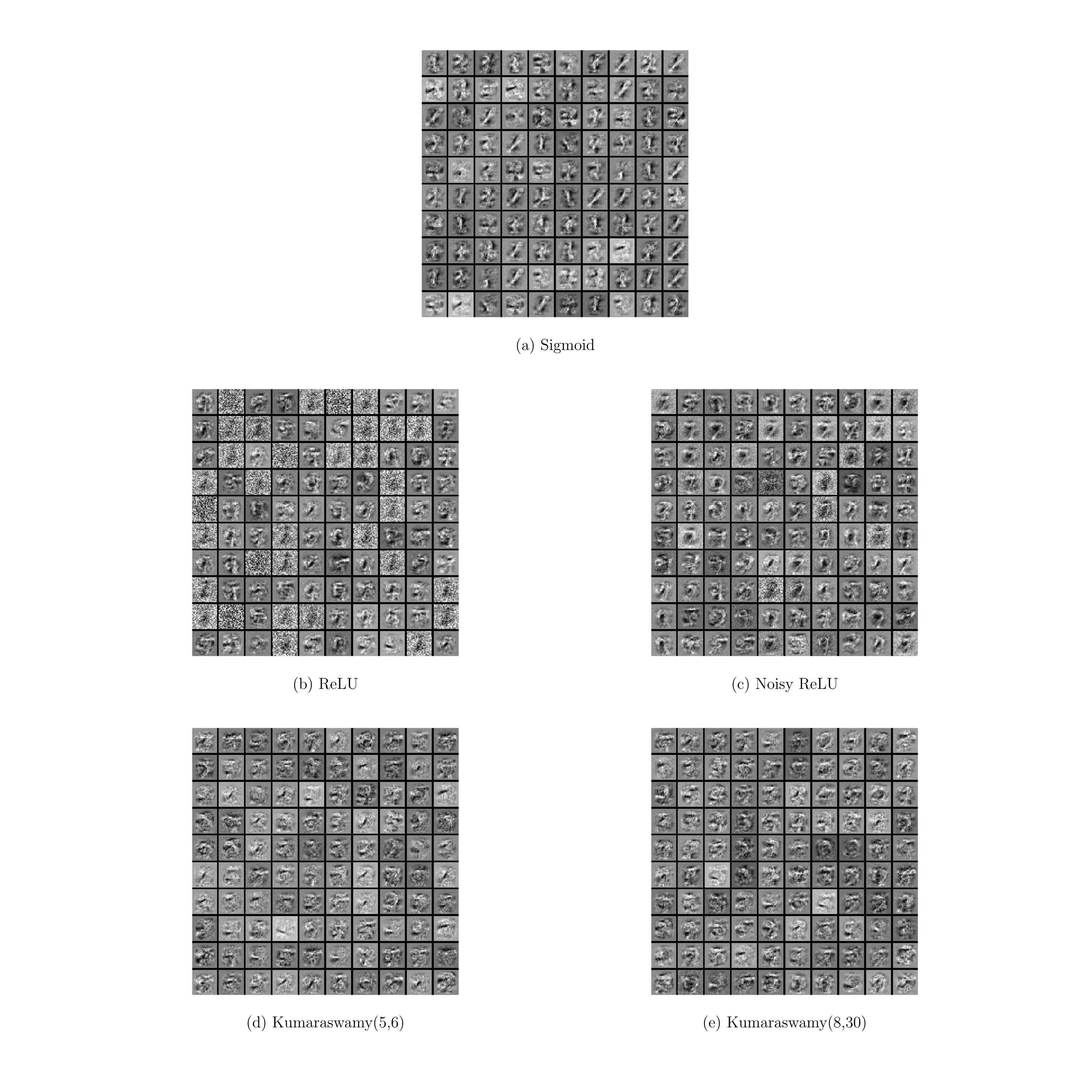}}
\caption{A visualization of first $100$ of the input-to-hidden weights (features). It is apparent that the patterns learned by the neural network with the considered in the paper units differ in shapes. As expected, application of ReLU results in saturation of several hidden units. The same outcome, but less evident, can be observed in the case of the Noisy ReLU. The Kumaraswamy unit do not lead to saturated or degenerated hidden units.}
\label{fig:weights}
\end{center}
\vskip -0.2in
\end{figure*}

\begin{figure*}[!htbp]
\vskip 0.2in
\begin{center}
\centerline{\includegraphics[width=1\textwidth]{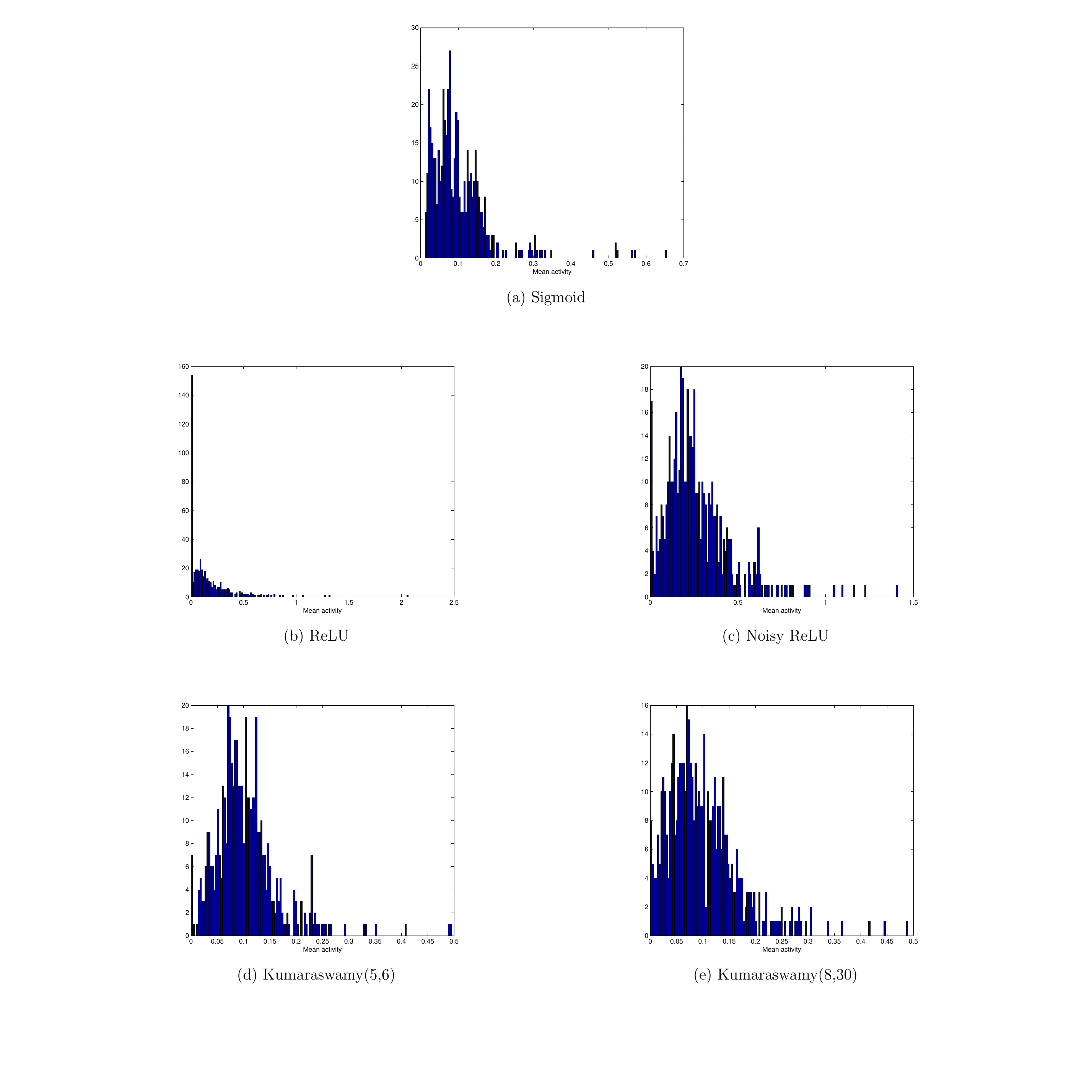}}
\caption{Mean activity of hidden units calculated on the test set. The sigmoid neural network has all hidden units with non-zero activity and some neurons are very active (above $0.5$). Application of ReLU results in many saturated hidden units while the Noisy ReLU alleviates this effect. Nonetheless, for ReLU and Noisy ReLU there are some hidden units with very strong response (i.e. above $1$). The utilization of Kumaraswamy units leads to a representation in which all neurons have activity less than $0.5$ on average.}
\label{fig:activity}
\end{center}
\vskip -0.2in
\end{figure*}

\paragraph{Results} The results for the considered activation non-linear activation functions are gathered in Table \ref{tab:results}. A single run of a training procedure for the considered units is presented in Figure \ref{fig:performance}. Moreover, in order to get better insight into the trained representation by the considered non-linearities input-to-hidden weights are depicted in Figure \ref{fig:weights} and the mean activities of hidden units are demonstrated in Figure \ref{fig:activity}.

\begin{table}[!htbp]
\centering
\caption{Test classification error and test cross-entropy loss for single-layer neural network with different hidden units on MNIST averaged over $5$ experiment runs. The best results are in bold.}
\vskip 5mm
\label{tab:results}
\resizebox{\columnwidth}{!}{
\begin{tabular}{l|c|c}
\hline
\textbf{Hidden unit} & \textbf{Error} [$\%$] & \textbf{Cross-Entropy}\\
\hline
sigmoid				& 5.85 & 0.21 \\
ReLU 				& 5.44 & 0.19 \\
Noisy ReLU 			& 5.79 & 0.21 \\
Kumaraswamy(5,6) 	& 5.18 & 0.17 \\
Kumaraswamy(8,30)  	& \textbf{4.87} & \textbf{0.16}\\
\end{tabular}}
\end{table}

\paragraph{Discussion} According to the results (see Table \ref{tab:results}) we can conclude that the Kumaraswamy unit indeed tends to be preferable in comparison to sigmoid unit, ReLU and Noisy ReLU. The Kumaraswamy unit obtained the best results in terms of the test classification error and the test cross-entropy loss. The comparison of the two considered possible values of the scale parameters reveals that the Kumaraswamy$(8,30)$ performs better than the Kumaraswamy$(5,6)$, i.e., the bunches of $30$ neurons with $8$ elements seem to be more suitable in training a neural network.

It is a well-known fact that application of ReLU can cause saturation of some hidden units. However, we notice that the saturation of several hidden units resulted in faster convergence of ReLU and Noisy ReLU in comparison to other activation non-linearities (see Figure \ref{fig:performance}). For instance, ReLU obtained the minimum after $18$ epochs while the sigmoid neural network converged after $65$ iterations. It is worth noticing that the Kumaraswamy unit obtained the the best result (better by about $3\%$ in comparison to others) just after $10$ epochs.

In Figure \ref{fig:weights} the learned features (input-to-hidden weights) for the considered types of hidden units are depicted. The features learned by the sigmoid neural network represent different patterns but some of them seem to be redundant, e.g., there are many patterns which look like diagonal stroke. On the contrary, ReLU and Noisy ReLU allow to obtain more diverse features. However, there are many neurons which are saturated, but in the case of Noisy ReLU this effect is less evident. Slightly different features in shape are obtained by Kumaraswamy units. These are more similar to the ones learned by ReLU, nevertheless, they are not saturated or degenerated as in the case of sigmoid units.

Next, we investigate the impact of different non-linearities on activity of hidden units. The histograms of mean activity of neurons measured on the test set given in Figure \ref{fig:activity} indicate that on average the ReLU, Noisy ReLU and Kumaraswamy units lead to larger number of neurons with activation smaller than $0.01$ in comparison to the sigmoid units, namely, ReLU: 150, Noisy ReLU: 19, Kumaraswamy$(5,6)$: 7, Kumaraswamy$(8,30)$: 8, sigmoid: 0. However, Kumaraswamy unit has two advantages over ReLU and Noisy ReLU. First, there are less saturated neurons. Second, all Kumaraswamy units have mean activity less than $0.5$ while in the case of ReLU and Noisy ReLU there are some units with very strong activation, i.e., above $0.5$ or even larger than $1$.

\section{Conclusion}
\label{sect:conclusion}

In this work we introduced a new idea of improving neural networks with bunch of neurons, i.e., replicas of the same neurons which consist of independent elements. The bunch of neurons can be easily modeled by the generalized Kumaraswamy distribution which resulted in a formulation of new non-linear activation function we refer to as \textit{Kumaraswamy unit}. A nice property of the Kumaraswamy unit is that for properly chosen parameters it can be approximately shaped as ReLU and it returns values between $0$ and $1$. In the experiment the performance of the neural network with the Kumaraswamy unit was compared with other activation functions, namely, sigmoid unit, ReLU and Noisy ReLU. The obtained results on MNIST seem to confirm the supremacy of the Kumaraswamy unit, nonetheless, this statement needs to be confirmed with more thorough studies. We believe that the performed experiment gives a good starting point for further research on the Kumaraswamy units applied to deep neural networks.

\section*{Acknowledgments} 
The research conducted by the author has been partially co-financed by the Ministry of Science and Higher Education, Republic of Poland, grant No. B40020/I32.


\end{document}